\documentclass[runningheads]{llncs}
\newcommand{\comment}[1]{}

\usepackage{graphicx}
\usepackage{amsmath,amssymb} 
\usepackage{color}
\usepackage[width=122mm,left=12mm,paperwidth=146mm,height=193mm,top=12mm,paperheight=217mm]{geometry}
\usepackage{hyperref}
\usepackage{xcolor}
\definecolor{orange}{cmyk}{0,0.6,1,0}
\definecolor{ForestGreen}{rgb}{0.0, 0.5, 0.0}

\begin{document}
\pagestyle{headings}
\mainmatter

\title{Full-body High-resolution Anime Generation with Progressive Structure-conditional Generative Adversarial Networks}

\titlerunning{Full-body High-resolution Anime Generation with PSGAN}

\authorrunning{Hamada \textit{et al.}}

\author{Koichi Hamada, 
Kentaro Tachibana,
Tianqi Li,\\ 
Hiroto Honda,
and Yusuke Uchida}

\institute{DeNA Co., Ltd., Tokyo, Japan}

\maketitle

\begin{abstract}
We propose Progressive Structure-conditional Generative Adversarial Networks (PSGAN), a new framework that can generate full-body and high-resolution character images based on structural information. Recent progress in generative adversarial networks with progressive training has made it possible to generate high-resolution images. However, existing approaches have limitations in achieving both high image quality and structural consistency at the same time. Our method tackles the limitations by progressively increasing the resolution of both generated images and structural conditions during training. In this paper, we empirically demonstrate the effectiveness of this method by showing the comparison with existing approaches and video generation results of diverse anime characters at 1024$\times$1024 based on target pose sequences. We also create a novel dataset containing full-body 1024$\times$1024 high-resolution images and exact 2D pose keypoints using Unity 3D Avatar models.

\keywords{Generative Adversarial Networks; Anime Generation; Image Generation; Video Generation}
\end{abstract}

\section{Introduction}
Recently automatic image and video generation using deep generative models has been studied \cite{Goodfellow+14,Karras+18,Vondrick+16}. These are useful for media creation tools such as photo editing, animation production and movie making.
Focusing on anime creation, automatic character generation can inspire experts to create new characters, and also can contribute to reducing costs for drawing animation.
Jin et al. \cite{Jin+17} focuses on image generation for anime character faces with GAN architecture. However full-body character generation has not been studied enough.
Generation of images for anime characters which only focused on face images was proposed,  however, its quality was not satisfactory for animation production requirements \cite{Jin+17}. 
To generate full-body characters automatically and add actions to them with high quality is a great help for making new characters and drawing animations.
Therefore, we work on generating full-body character images and adding actions to them (i.e., video generation) with high quality.

There remain two problems to applying full-body character generation to animation production: (i) generation with high-resolution, (ii) generation with specified sequences of poses. 

Generative Adversarial Networks (GANs) \cite{Goodfellow+14} are one of the most promising candidates as a framework applied to a diverse range of image generation tasks \cite{Jin+17,Radford+16,Reed+16,Isola+17,Zhu+17,Ma+17}. Recent progress of GANs with hierarchical and progressive structures has been realizing high-resolution and high-quality image generation \cite{Karras+18}, text-to-image synthesis \cite{Zhang+18a,Zhang+18b}, and image synthesis from label map \cite{Wang+18}. However, It is still a challenge for GANs to generate structured objects consistent with global structures \cite{Goodfellow17}, such as full-body character generation. On the other hand, GANs with structural conditions, such as pose keypoints and facial landmarks, have been also proposed \cite{Ma+17,Ma+18,Balakrishnan+18,Siarohin+18,Si+18,Hu+18,Qiao+18}. However, their image resolution and quality are insufficient.

\begin{figure}[t!]
\centering
\includegraphics[width=\linewidth]{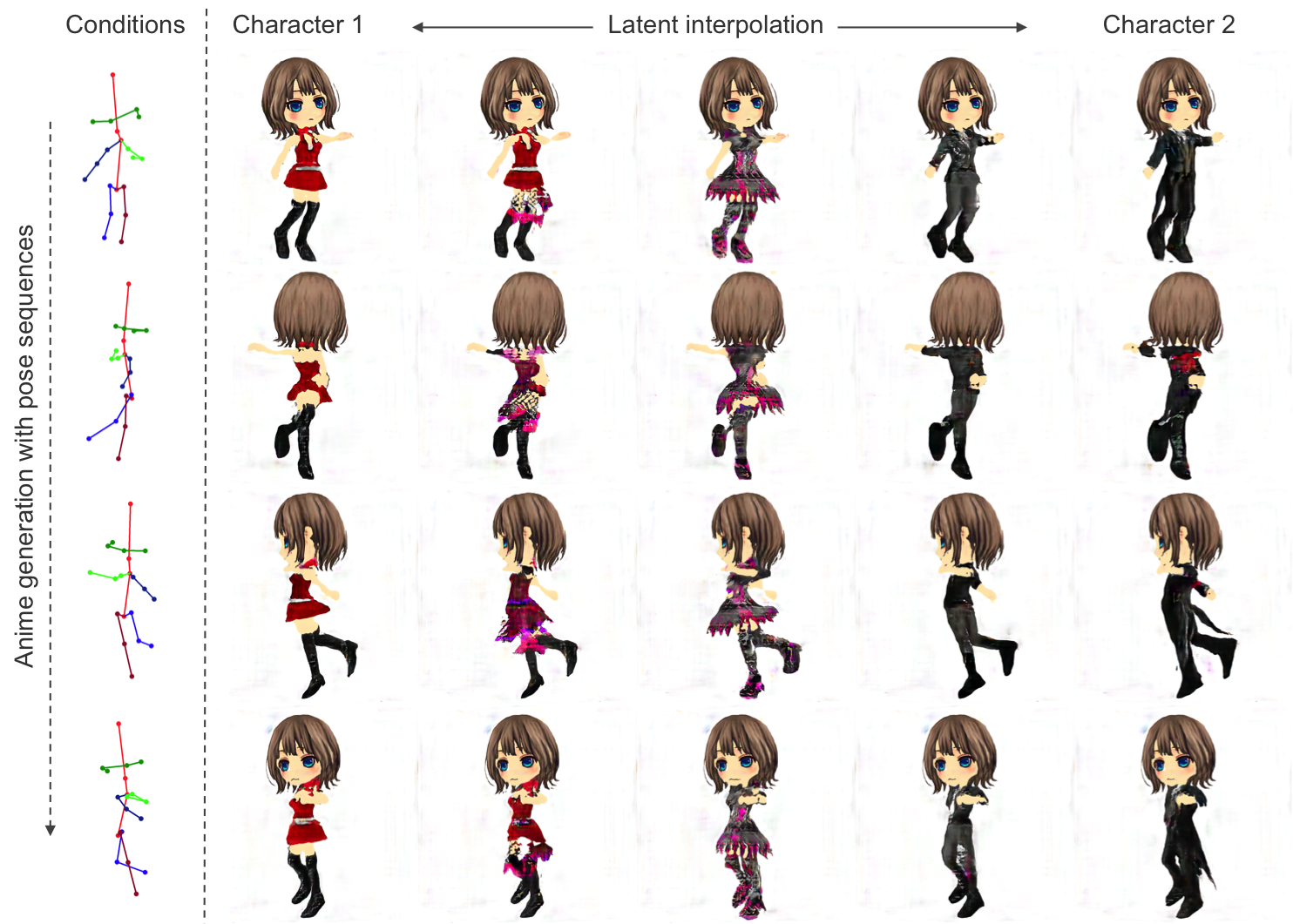}
\vspace{-6mm}
\caption{Generated images of full-body anime characters at 1024$\times$1024 by PSGAN with a test pose sequence. A generated anime at 1024$\times$1024 by PSGAN is at \url{https://youtu.be/bIi5gSITK0E}.}
\vspace{-6mm}

\label{fig:generation_character}
\end{figure}

We propose Progressive Structure-conditional GANs (PSGAN) to tackle these problems by imposing the structural conditions at each scale generation with progressive training. We show that PSGAN is able to generate full body anime characters and animations with target pose sequences at 1024$\times$1024 resolution. As PSGAN generates images with latent variables and structural conditions, PSGAN is able to generate controllable animations for various characters with target pose sequences. Fig.~\ref{fig:generation_character} shows some example of animation generation results. 

\section{Proposed Methods}
\subsection{Progressive Structure-conditional GANs}
\begin{figure}[tb]
\centering
\includegraphics[width=10.3cm]{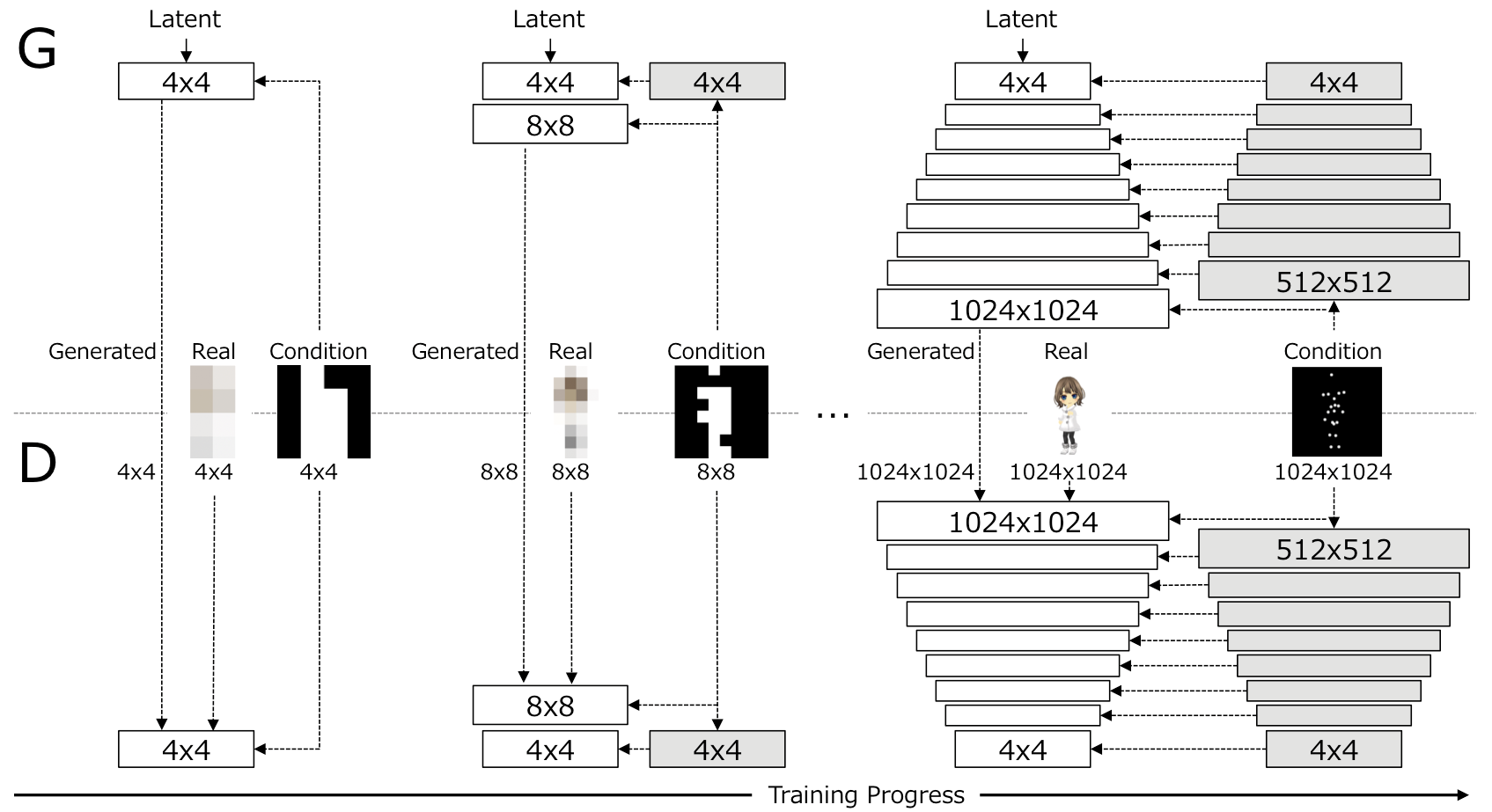}
\vspace{-3mm}
\caption{Generator (G) and Discriminator (D) architecture of PSGAN.}
\vspace{-5mm}
\label{fig:model}
\end{figure}

Our key idea is to learn image representation with structural conditions progressively. Fig.~\ref{fig:model} shows generator $G$ and discriminator $D$ architecture of PSGAN. PSGAN increases the resolution of generated images with structural conditions at each scale and generates high-resolution images. We adopt the same architecture of the image generator and discriminator as Progressive GAN \cite{Karras+18}, except that we impose structural conditions on both the generator and discriminator at each scale by adding pose maps with corresponding resolutions, which significantly stabilizes training. GANs with structural conditions have also been proposed \cite{Ma+17,Ma+18,Balakrishnan+18,Siarohin+18,Si+18,Hu+18,Qiao+18}. They exploit a single-scale condition while we use multi-scale conditions. 
More specifically, we downsample the full-resolution structural condition map at each scale to form multi-scale condition maps.
For each scale, the generator generates an image from a latent variable with a structural condition and the discriminator discriminates the generated images and real images based on the structural conditions. N$\times$N white boxes stand for learnable convolution layers operating on N$\times$N spatial resolution. N$\times$N gray boxes stand for non-learnable downsampling layers for structural conditions, which reduce spatial resolution of the structural condition map to N$\times$N. We use M channels for representation of M-dimensional structural conditions (e.g. M keypoints). 

\vspace{-2mm}
\subsection{Automatic Dataset Construction with Exact Pose Keypoints from Unity 3D Models}

We create a novel dataset containing full-body high-resolution anime character images and exact 2D pose keypoints using the Unity\footnote{Unity: \url{https://unity3d.com}} 3D models for various poses, in a similar manner as is done in \cite{Chen+16,Varol+17} for photo-realistic images. We use various motions and costumes of full-body character models to create this dataset. The four key features of our methodology are the following:
1) Pose Diversity: To generate smooth and natural animation we prepare a very wide variety of pose conditions. We generate high-resolution images and pose keypoint coordinates of various poses for reproducing smooth and natural continuous motion by capturing images and exactly calculating the coordinates while each Unity 3D model is moving with each Unity motion. 
2) Exact pose keypoints: Direct calculation of pose keypoint coordinates from the Unity model makes it possible to calculate the coordinates with no estimation error. 
3) Infinite number of training images: An infinite number of synthetic images with keypoint maps are obtained by generating 3D modeled avatars using Unity automatically. Various images with keypoints can be created by replacing detachable items for each Unity 3D model.
4) Background elimination: We can set the background color to white and erase unnecessary information to avoid negative effects on image generation.

\section{Experiments}
We evaluate the effectiveness of the proposed method in terms of quality and structural consistency of generated images on the Avatar Anime-Character dataset and DeepFashion dataset. We show comparisons between our method and existing works.

\vspace{-2mm}
\subsection{Datasets}
In this section, we describe our dataset preparation methodology. For PSGAN we require pairs of image and keypoint coordinates. We prepare the original Avatar Anime-Character dataset synthesized by Unity, and DeepFashion dataset \cite{Liu+16} with keypoints detected by Openpose \cite{Cao+16}. 

\begin{figure}[t!]
\centering
\includegraphics[width=11.5cm]{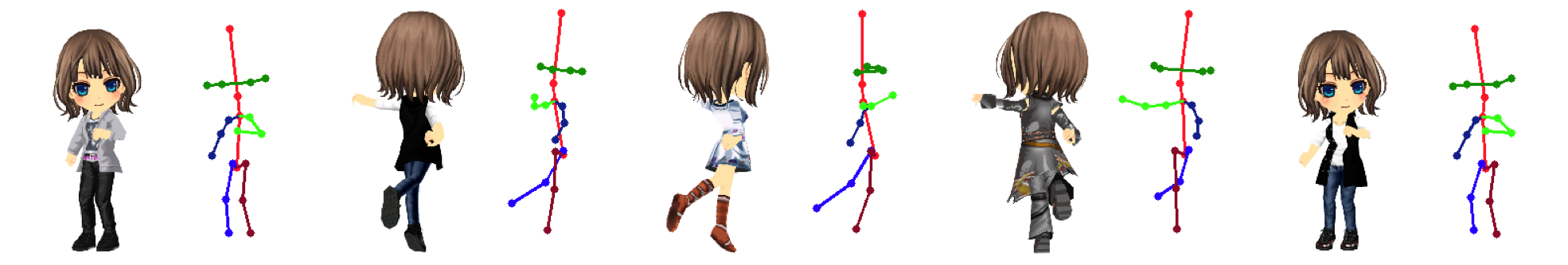}
\vspace{-3mm}
\caption{Samples of Avatar Anime-Character dataset.}
\vspace{-5mm}

\label{fig:dataset}
\end{figure}

\vspace{1mm}
\noindent{\bf Avatar Anime-Character Dataset.} 
We create a dataset of full-body 1024$\times$1024 high-resolution images with exact 2D pose keypoints from Unity 3D Avatar models based on the above proposed method. We divide several continuous actions of one avatar into 600 poses, and calculate keypoints in each pose. We conduct such process for 69 kinds of costumes, and obtain 47,400 images in total. We also obtain 20 keypoints based on the location of the bones of the 3D model. Fig.~\ref{fig:dataset} shows samples of created data. Anime characters (left of pair) and pose images (right of pair) are shown.

\vspace{1mm}
\noindent{\bf DeepFashion Dataset.} 
The DeepFashion dataset (In-shop Clothes Retrieval Benchmark) \cite{Liu+16} consists of 52,712 in-shop clothes images, and 200,000 cross-pose/scale pairs. All images are in 256$\times$256 resolution and are richly annotated by bounding box, clothing type and pose type. However, none of them has keypoint annotations. Following \cite{Ma+17}, we use Openpose \cite{Cao+16} to extract keypoint coordinates from images. The number of keypoints is 18 and examples with less than 10 detected keypoints are omitted.

\vspace{-2mm}
\subsection{Experimental Setups}
We use the same stage design and the same loss function as \cite{Karras+18}. We train networks with 600k images and structural conditions for each stage and use WGAN-GP loss \cite{Gulrajani+17} with $n_{critic} = 1$. We use a minibatch size 16 at the stage for 4$\times$4 - 128$\times$128 image generation and gradually decrease it to 12 for 256$\times$256, 5 for 512$\times$512, and 2 for 1024$\times$1024 respectively due to GPU memory constraints. 
We use M channels of structural conditions as pose keypoints. M is 20 for the Avatar Anime-Character dataset and 18 for DeepFashion dataset. At each scale, the single pixel value at the corresponding keypoint coordinate is set to 1 and -1 elsewhere. For downsampling the condition map, we use max-pooling with kernel size 2 and stride 2 at each scale. We train the networks using Adam \cite{Kingma+15} with $\beta_{1}=0$, $\beta_{2}=0.99$. We use $\alpha=0.001$ at the stage for 4$\times$4 - 64$\times$64 image generation and gradually decrease it to $\alpha=0.0008$ for 128$\times$128, $\alpha=0.0006$ for 256$\times$256, $\alpha=0.0002$ for 512$\times$512, and $\alpha=0.0001$ for 1024$\times$1024 respectively. 

\vspace{-2mm}
\subsection{Avatar Anime-Character Generation at 1024$\times$1024}
We show examples of a variety of anime characters and animations generated at 1024$\times$1024 by PSGAN.
Fig.~\ref{fig:generation_character} shows generated results of full-body anime characters at 1024$\times$1024 with a test pose sequence. We can generate new full-body anime characters by interpolating latent variables corresponding to anime characters with different costumes (character 1 and 2) for various poses. By fixing the latent variables and giving continuous pose sequences to the network, we can generate an animation of the specified anime characters\footnote{An illustration video for adding action to full-body anime characters with PSGAN is at \url{https://youtu.be/0LQlfkvQ3Ok}}.

\vspace{-2mm}
\subsection{Comparison of PSGAN, Progressive GAN, and PG2}
\begin{figure}[t!]
\centering
\includegraphics[width=11.4cm]{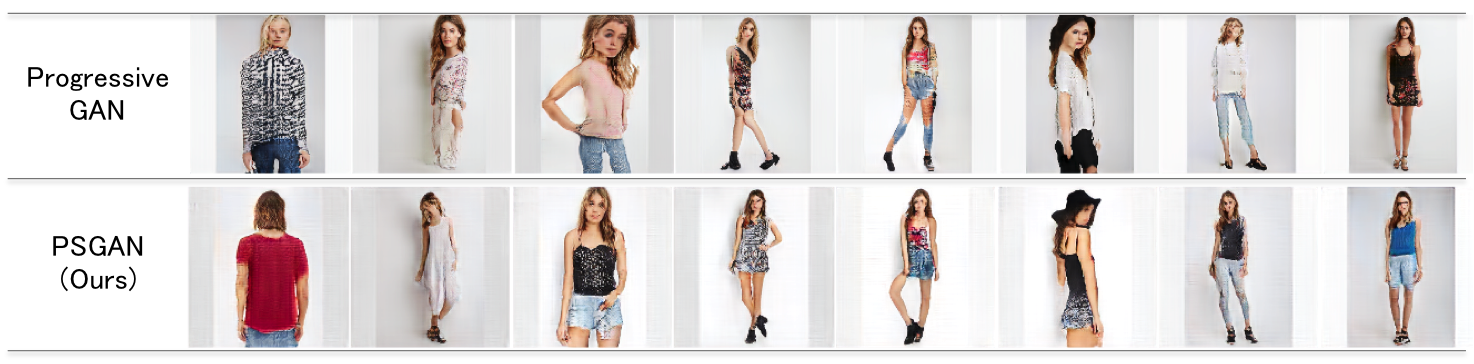}
\vspace{-3mm}
\caption{Comparison of structural consistency with \cite{Karras+18} on DeepFashion dataset.}
\vspace{-3mm}
\label{fig:comparison_pggan}

\end{figure}

First, we evaluate structural consistency of PSGAN compared to Progressive GAN \cite{Karras+18}. Fig.~\ref{fig:comparison_pggan} shows generated images on the DeepFashion dataset (256$\times$256) by Progressive GAN and PSGAN. We observe that Progressive GAN is not capable of generating natural images consistent with their global structures (for example, left four images). On the other hand, PSGAN can generate plausible images consistent with their global structures by imposing the structural conditions at each scale. 

\begin{figure}[t!]
\centering
\includegraphics[width=\linewidth]{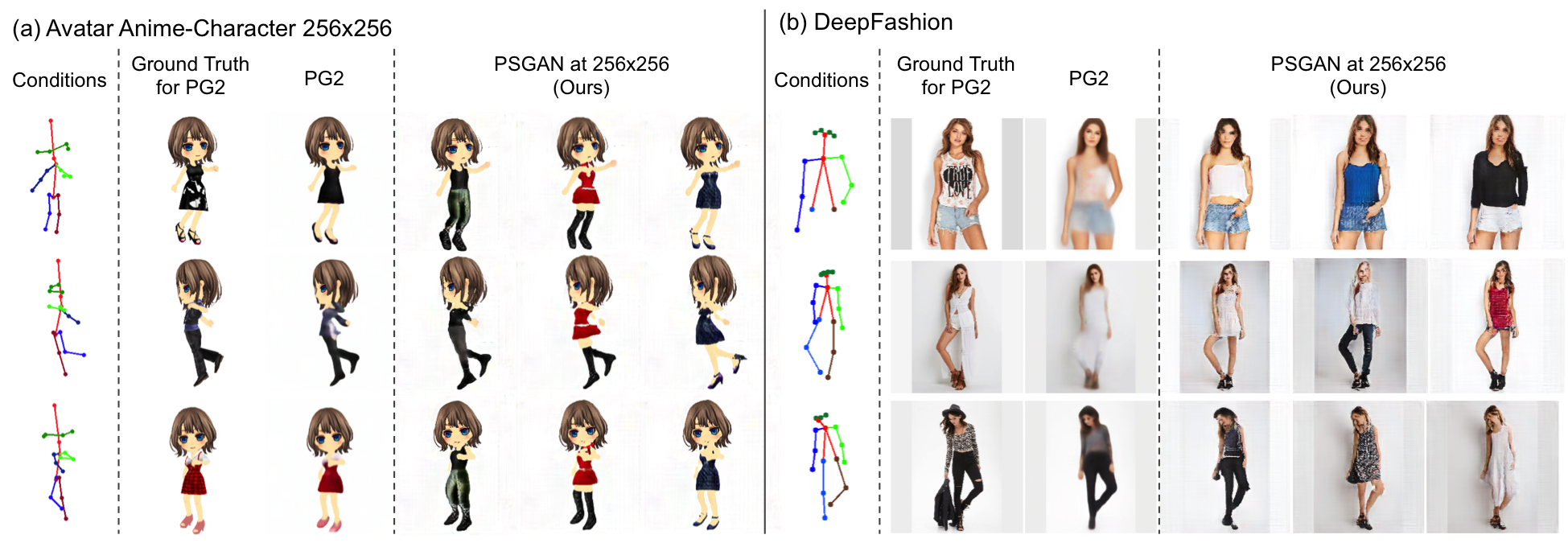}
\vspace{-7mm}
\caption{Comparison of generated image quality based on pose conditions with \cite{Ma+17} at 256$\times$256 on (a) Avatar Anime-Character dataset and (b) DeepFashion dataset.}
\vspace{-5mm}

\label{fig:comparison_pg2}
\end{figure}

Next, we evaluate image quality on pose conditional image generation of PSGAN compared to Pose Guided Person Image Generation (PG2) \cite{Ma+17}. PG2 requires a source image and a corresponding target pose to convert the source image to an image with the structure of the target pose. Meanwhile, PSGAN generates an image with the structure of the target pose from latent variables and the target pose and does not need paired training images. Fig.~\ref{fig:comparison_pg2} shows generated images of PSGAN and PG2 on the 256$\times$256 resolution version of the Avatar dataset and DeepFashion dataset. We pick the weight parameter for L1 loss of PG2 (which affects image quality) to 1.0. The input image of PG2 is omitted. We can observe the generated images of PSGAN are less blurry and more detailed than PG2 due to structural conditions imposed at each scale. 

\section{Conclusion}
In this paper, we have demonstrated smooth and high-resolution animation generation with PSGAN. We have shown that the method can generate full-body anime characters and the animations based on target pose sequences at 1024$\times$1024 resolution. 
PSGAN progressively increases the resolution of generated images with structural conditions at each scale during training and generates detailed images for structured objects, such as full-body characters. As PSGAN generates images with latent variables and structural conditions, it is able to generate controllable animations with target pose sequences. 
Our experimental results demonstrate that PSGAN can generate a variety of high-quality anime characters from random latent variables, and smooth animations by imposing continuous pose sequences as structural conditions. Since the experimental setting still remains limited, such as one avatar and several actions, we plan to conduct experiments and evaluation in various conditions. We plan to make the Avatar Anime-Character dataset available in the near future.

\clearpage

\bibliographystyle{splncs04}
\bibliography{egbib}

\begin{thebibliography}{10}
\providecommand{\url}[1]{\texttt{#1}}
\providecommand{\urlprefix}{URL }
\providecommand{\doi}[1]{https://doi.org/#1}

\bibitem{Balakrishnan+18}
Balakrishnan, G., Zhao, A., Dalca, A.V., Durand, F., Guttag, J.: Synthesizing
  images of humans in unseen poses. In: Proc. of CVPR (2018)

\bibitem{Cao+16}
Cao, Z., Simon, T., Wei, S.E., Sheikh, Y.: Realtime multi-person 2d pose
  estimation using part affinity fields. In: Proc. of CVPR (2016)

\bibitem{Chen+16}
Chen, W., Wang, H., Li, Y., Su, H., Wang, Z., Tu, C., Lischinski, D., nad
  Baoquan~Chen, D.C.O.: Synthesizing training images for boosting human 3d pose
  estimation. In: Proc. of 3D Vision (2016)

\bibitem{Goodfellow17}
Goodfellow, I.: {NIPS} 2016 tutorial: Generative adversarial networks.
  arXiv:1701.00160  (2017)

\bibitem{Goodfellow+14}
Goodfellow, I.J., Pouget-Abadie, J., Mirza, M., Xu, B., Warde-Farley, D.,
  Ozair, S., Courville, A., Bengio, Y.: Generative adversarial nets. In: Proc.
  of NIPS (2014)

\bibitem{Gulrajani+17}
Gulrajani, I., Ahmed, F., Arjovsky, M., Dumoulin, V., Courville, A.: Improved
  training of wasserstein gans. In: Proc. of NIPS (2017)

\bibitem{Hu+18}
Hu, Y., Wu, X., Yu, B., He, R., Sun, Z.: Pose-guided photorealistic face
  rotation. In: Proc. of CVPR (2018)

\bibitem{Isola+17}
Isola, P., Zhu, J.Y., Zhou, T., Efros, A.A.: Image-to-image translation with
  conditional adversarial networks. In: Proc. of CVPR (2017)

\bibitem{Jin+17}
Jin, Y., Zhang, J., Li, M., Tian, Y., Zhu, H.: Towards the high-quality anime
  characters generation with generative adversarial networks. In: Proc. of NIPS
  Workshop on Machine Learning for Creativity and Design (2017)

\bibitem{Karras+18}
Karras, T., Aila, T., Laine, S., Lehtinen, J.: Progressive growing of gans for
  improved quality, and stability, and variation. In: Proc. of ICLR (2018)

\bibitem{Kingma+15}
Kingma, D.P., Ba, J.: Adam: A method for stochastic optimizations. In: Proc. of
  ICLR (2015)

\bibitem{Liu+16}
Liu, Z., Luo, P., Qiu, S., Wang, X., Tang, X.: {DeepFashion}: Powering robust
  clothes recognition and retrieval with rich annotations. In: Proc. of CVPR
  (2016)

\bibitem{Ma+18}
Ma, L., Sun, Q., Georgoulis, S., Gool, L.V., Schiele, B., Fritz, M.:
  Disentangled person image generation. In: Proc. of CVPR (2018)

\bibitem{Ma+17}
Ma, L., Sun, Q., Jia, X., Schiele, B., Tuytelaars, T., Gool, L.V.: Pose guided
  person image generation. In: Proc. of NIPS (2017)

\bibitem{Qiao+18}
Qiao, F., Yao, N., Jiao, Z., Li, Z., Chen, H., Wang, H.: Geometry-contrastive
  generative adversarial network for facial expression synthesis.
  arXiv:1802.01822  (2018)

\bibitem{Radford+16}
Radford, A., Metz, L., Chintala, S.: Unsupervised representation learning with
  deep convolutional generative adversarial networks. In: Proc. of ICLR (2016)

\bibitem{Reed+16}
Reed, S., Akata, Z., Yan, X., Logeswaran, L., Schiele, B., Lee, H.: Generative
  adversarial text to image synthesis. In: Proc. of ICML (2017)

\bibitem{Si+18}
Si, C., Wang, W., Wang, L., Tan, T.: Multistage adversarial losses for
  pose-based human image synthesis. In: Proc. of CVPR (2018)

\bibitem{Siarohin+18}
Siarohin, A., Sangineto, E., Lathuiliere, S., Sebe, N.: Deformable gans for
  pose-based human image generation. In: Proc. of CVPR (2018)

\bibitem{Varol+17}
Varol, G., Romero, J., Martin, X., Mahmood, N., Black, M.J., Laptev, I.,
  Schmid, C.: Learning from synthetic humans. In: Proc. of CVPR (2017)

\bibitem{Vondrick+16}
Vondrick, C., Pirsiavash, H., Torralba, A.: Generating videos with scene
  dynamics. In: Proc. of NIPS (2016)

\bibitem{Wang+18}
Wang, T.C., Liu, M.Y., Zhu, J.Y., Tao, A., Kautz, J., Catanzaro, B.:
  High-resolution image synthesis and semantic manipulation with conditional
  gans. In: Proc. of CVPR (2018)

\bibitem{Zhang+18b}
Zhang, H., Xu, T., Li, H., Zhang, S., Wang, X., Huang, X., Metaxas, D.:
  Stackgan++: Realistic image synthesis with stacked generative adversarial
  networks. TPAMI  (2018)

\bibitem{Zhang+18a}
Zhang, Z., Xie, Y., Yang, L.: Photographic text-to-image synthesis with a
  hierarchically-nested adversarial network. In: Proc. of CVPR (2018)

\bibitem{Zhu+17}
Zhu, J.Y., Park, T., Isola, P., Efros, A.A.: Unpaired image-to-image
  translation using cycle-consistent adversarial networks. In: Proc. of ICCV
  (2017)

\end{thebibliography}

\clearpage

\title{Supplementary Material: \\
Full-body High-resolution Anime Generation with Progressive Structure-conditional Generative Adversarial Networks}

\titlerunning{Full-body High-resolution Anime Generation with PSGAN}

\authorrunning{Hamada \textit{et al.}}

\author{Koichi Hamada, 
Kentaro Tachibana,
Tianqi Li,\\ 
Hiroto Honda,
and Yusuke Uchida}

\institute{DeNA Co., Ltd., Tokyo, Japan}
\maketitle 
\section{Diversity of generated images based on pose conditions}

\begin{figure}
\centering
\includegraphics[width=\linewidth]{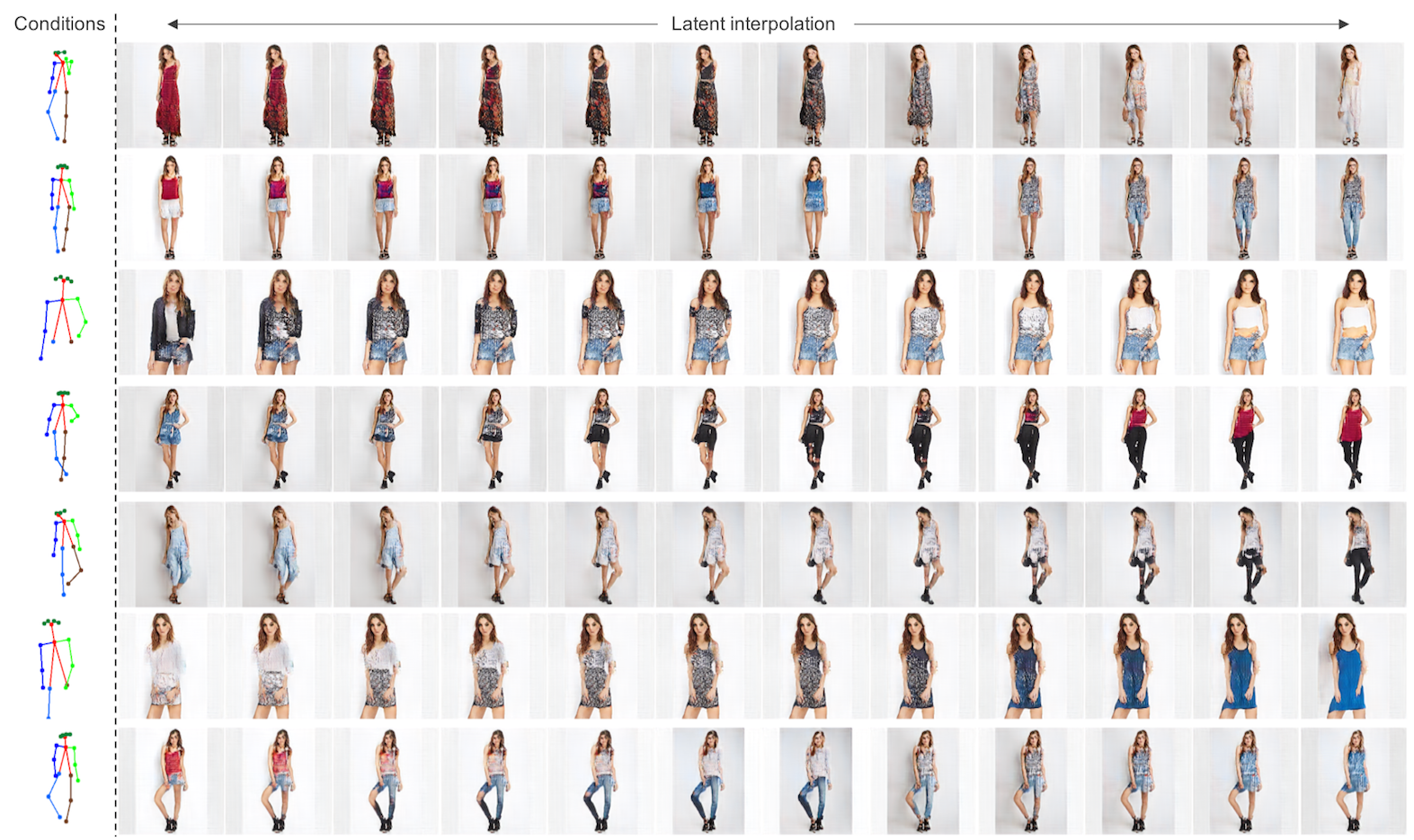}
\vspace{-6mm}
\caption{Generated images of PSGAN with latent interpolations on the DeepFashion dataset at 256$\times$256.}
\label{fig:deepfashion_interpolation}
\end{figure}

Fig.~\ref{fig:deepfashion_interpolation} shows generated images of PSGAN with latent interpolations on the DeepFashion dataset. PSGAN can generate a diverse set of images for a wide variety of poses. We can also generate new clothes by interpolating latent variables corresponding to different clothes for each given pose.

\end{document}